\documentclass{article}

\usepackage[cloudcomputing]{cosparws_abs} %uncomment for Cloud Computing Workshop

\usepackage{longtable}

\usepackage{subcaption}
\usepackage{graphicx}
\usepackage[utf8]{inputenc} % allow utf-8 input
\usepackage[T1]{fontenc}    % use 8-bit T1 fonts
\usepackage{hyperref}       % hyperlinks
\usepackage{url}            % simple URL typesetting
\usepackage{booktabs}       % professional-quality tables
\usepackage{amsfonts}       % blackboard math symbols
\usepackage{nicefrac}       % compact symbols for 1/2, etc.
\usepackage{microtype}      % microtypography

\title{Scalable Reverse Image Search Engine for NASA Worldview}

\usepackage{xcolor}

\newcommand*\samethanks[1][\value{footnote}]{\footnotemark[#1]}

\author{%
  Abhigya Sodani \thanks{Corresponding Author: \texttt{abhigyasodani@ucla.edu}}  \thanks{Work done as a researcher at Space ML\cite{LearningsfromFrontierDevelopmentLab}}  \qquad
  Michael Levy\samethanks{} \\

Anirudh Koul \qquad
  Meher Anand Kasam \qquad
  Siddha Ganju \qquad
}

\begin{document}

\maketitle

%\begin{abstract}
%  The abstract paragraph should be indented \nicefrac{1}{2}~inch (3~picas) on
%  both the left- and right-hand margins. Use 10~point type, with a vertical
%  spacing (leading) of 11~points.  The word \textbf{Abstract} must be centered,
%  bold, and in point size 12. Two line spaces precede the abstract. The abstract
%  must be limited to one paragraph.
%\end{abstract}

\section*{Abstract}

Earth science researchers often spend weeks sifting through decades of unlabeled satellite imagery (on NASA Worldview) in order to develop datasets on which they can start conducting research \cite{seeley2020kdf}. We developed an interactive, scalable and fast image similarity search engine (which can take one or more images as the query image) that automatically sifts through the unlabeled dataset reducing dataset generation time from weeks to minutes. In this work, we describe key components of the end to end pipeline.

Our similarity search system was created to be able to identify similar images from a potentially petabyte scale database that are similar to an input image, and for this we had to break down each query image into its features, which were generated by a classification layer stripped CNN trained in a supervised manner. To store and search these features efficiently, we had to make several scalability improvements. 

To improve the speed, reduce the storage, and shrink memory requirements for embedding search, we add a fully connected layer to our CNN make all images into a 128 length vector before entering the classification layers. This helped us compress the size of our image features from 2048 (for ResNet, which was initially tried as our featurizer) to 128 for our new custom model. Additionally, we utilize existing approximate nearest neighbor search libraries to significantly speed up embedding search. Our system currently searches over our entire database of images at 5 seconds per query on a single virtual machine in the cloud. In the future, we would like to incorporate a SimCLR based featurizing model which could be trained without any labelling by a human (since the classification aspect of the model is irrelevant to this use case). 

We also built a UX that enhanced Worldview with a human-in-the-loop interface that allows multi-stage refinement of query images. A user utilizes our image snipping tool on Worldview to choose a query area. Then, the search algorithm displays similar images in a tiled interface. The user selects images relevant to their particular task thereby refining the search results. This process is carried out iteratively until the entire dataset is collected.

\section{Introduction}
Worldview is a tool developed by NASA's Earth Observing System Data and Information System (EOSDIS) division. It allows users to interact with NASA satellite imagery superimposed onto a map of the Earth. 900 layers of this imagery are present on this application, and all these layers are updated daily. Worldview is powered by the NASA GIBS API which enables retrieval of satellite imagery at a certain location and zoom level with respect to the Earth's surface. 

The immense amount of image data contained by Worldview becomes an issue when no framework exists to search that data. An example user workflow that would encounter roadblocks due to this lack of search ability would be the creation of an image dataset of hurricanes. Currently,  such imagery must be procured by visually observing decades worth of satellite imagery. By using machine learning and approximate nearest neighbor techniques, we describe a solution to this problem in this paper. 

The problem of image searching has two main approaches: by metadata or by image features. In the metadata approach, each image is tagged with metadata of what it contains. When a user attempts to search for images (either via text or another image as input) the images with the metadata that most match query metadata are returned to the user. The limiting factor in the metadata approach is that generation of metadata would have to be done by hand since attempting to train a model that incorporated all "classes" of satellite imagery at a granular level would be a nearly impossible task. 

The second approach is to search by the image features themselves. In this approach, a trained CNN is utilized to convert each image into a list of integers, which we can call the images "features" because these integers represent the distinguishing features of an image. We then store these image features, and when provided a query image, use our trained CNN to convert the query image into its features. Then the query features are used to search our database for the stored images with closest matching features. The advantages to this approach lie in its ability to search images at a much from specific level than what metadata allows for (i.e a certain \textit{type} of hurricane) and that there becomes no need for a human to have to work with each image and tag it with metadata. 

For the feature-based approach to be time-feasible we need fast nearest neighbor searching upon the image features. There are several libraries that accomplish this approximation, the one we decided to use in this paper was the Annoy library from Spotify.   

\section{Related Work}
Much of the related work that this paper builds off of was done in creating the Approximate Nearest Neighbor search library Annoy \cite{ANN}. Annoy creates a binary tree forest of data point partitions split by a series of hyperplanes. Each node in the binary tree represents a hyperplane, and each leaf node represents the number of points in that given partition. When we search for the nearest neighbors for a given input, we simply determine which side of a given hyperplane node our input node is on, and then go to the node on that side of the hyperplane. This continues until we reach leaf nodes, which are all partitions that contain points from the original subspace. The leaf node partitions represent the set of points that are the nearest neighbors of the input point. To expand our set of nearest neighbors we can make a forest of hyperplane-divided binary trees, and all trees can be traversed simultaneously to find any number of N nearest neighbors. This method of fast, approximate searching was necessary to search efficiently through the 128-dimension feature set that represents every image in the dataset. 

\section{Pipeline}

Our reverse image search engine was based upon 52k images taken from NASA GIBS at zoom level 8 and 200 images taken from NASA GIBS at zoom level 4. Instead of storing these images in their entirety, we stored their distinguishing features as 128-length vectors. These distinguishing features were extracted by the vector created by the trained CNN right before the classifying layers. This allowed us to reduce the feature vector size from 2048 to 128. These vectors were stored in .ann (Annoy) files which could be quickly searched by the approximate nearest neighbor Annoy python library. The combination of these compressed features and Annoy search optimized both our database size and our search time.

The Annoy search returns the indexes of the features that most match the query image by the user. These indexes can the be converted into a NASA GIBS URL pointing to the image that the features represent. The URLS are then returned to the user. 

\begin{figure}[htp]
    \centering
    \includegraphics[width=10cm]{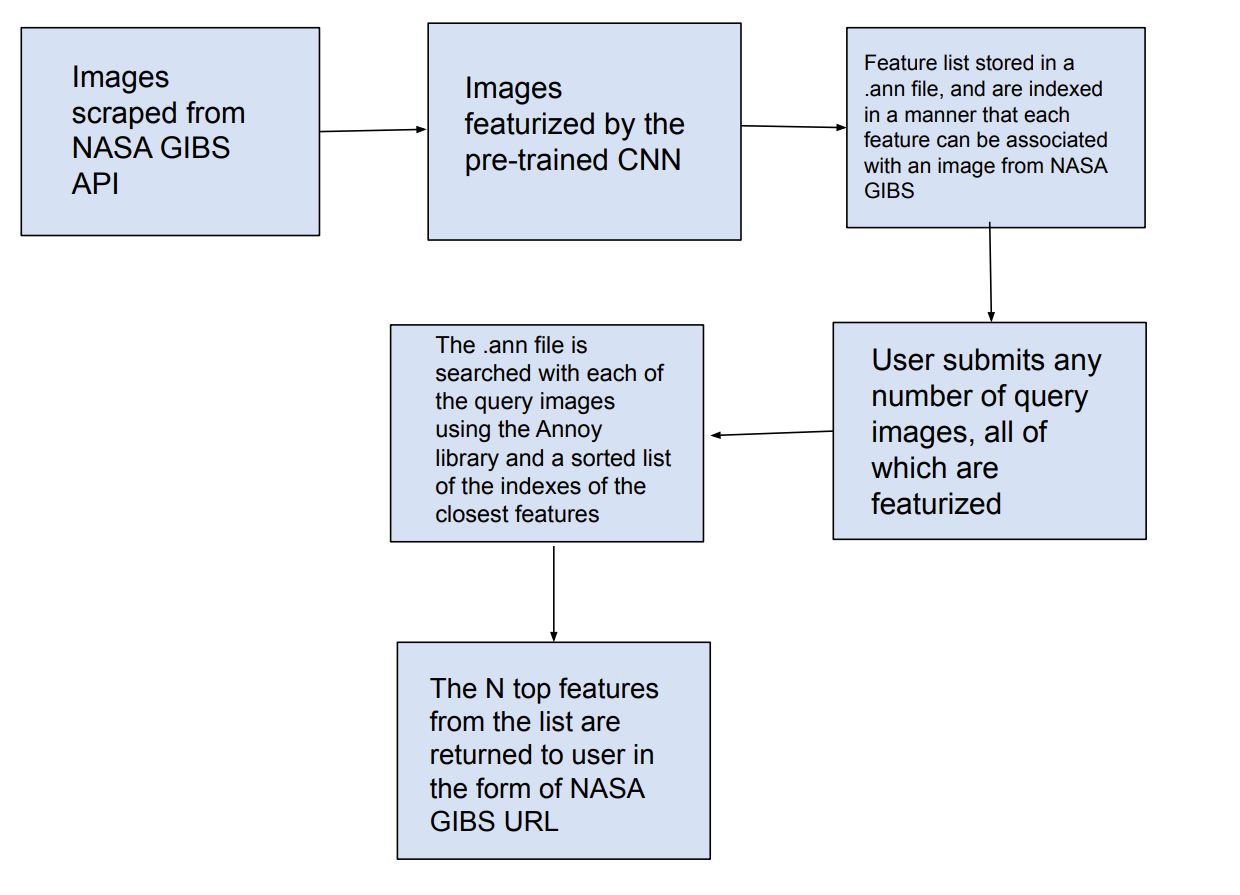}
    \caption{Pipeline of the Reverse Image Search}
    \label{fig:pipeline}
\end{figure}

We deployed this search engine system on a cloud VM where it ran in a Flask app. We then had a cloned NASA Worldview frontend access this Flask app, which would return the URLs of the closest images to a query image (see User Flow)

\section{Users Flow of Utilization of the Scalable Reverse Image Search Engine}
Here we provide sample screenshots that describes a typical users workflow and how the intended user uses the newly developed tool.

\begin{figure}[htp]
    \centering
    \includegraphics[width=15cm]{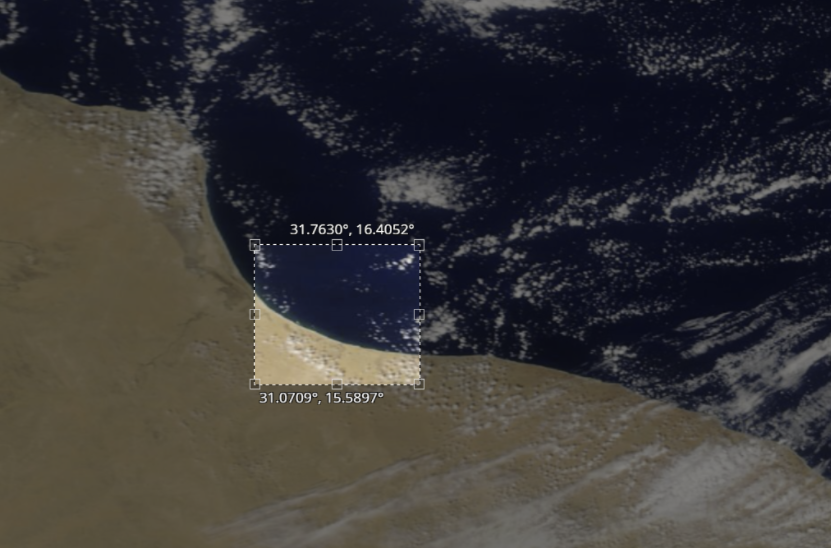}
    \caption{Step 1: The user snips an image of the region of interest}
    \label{fig:step1}
\end{figure}

\begin{figure}[htp]
    \centering
    \includegraphics[width=15cm]{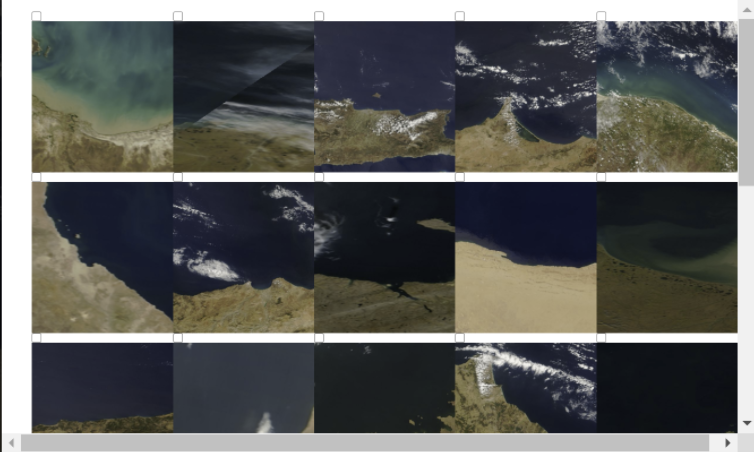}
    \caption{Step 2: The system returns a candidate list of similar regions of interest}
    \label{fig:step1}
\end{figure}

\begin{figure}[htp]
    \centering
    \includegraphics[width=15cm]{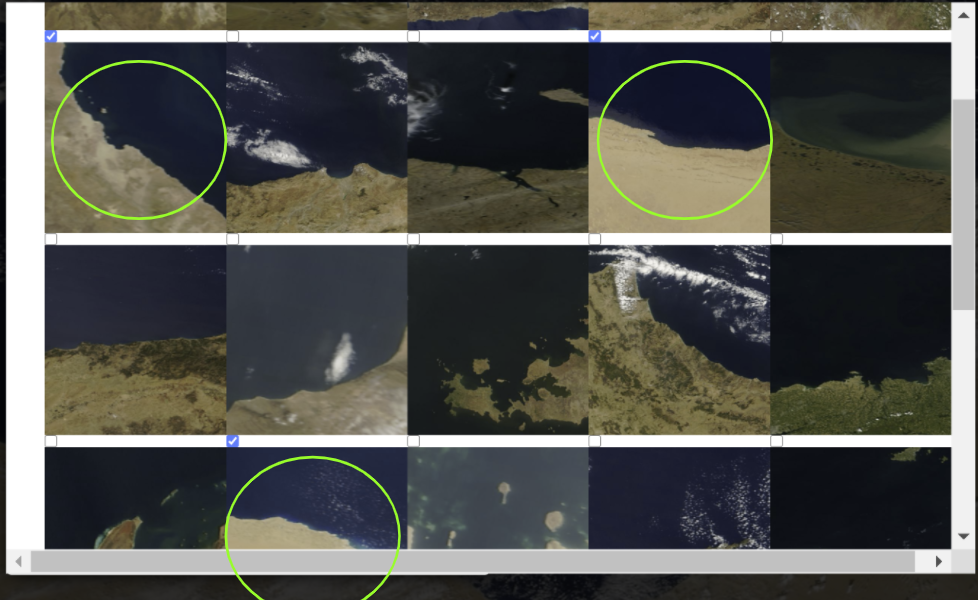}
    \caption{Step 3: The user selects more images of interest and the system refines the search results accordingly}
    \label{fig:step1}
\end{figure}

\begin{figure}[htp]
    \centering
    \includegraphics[width=15cm]{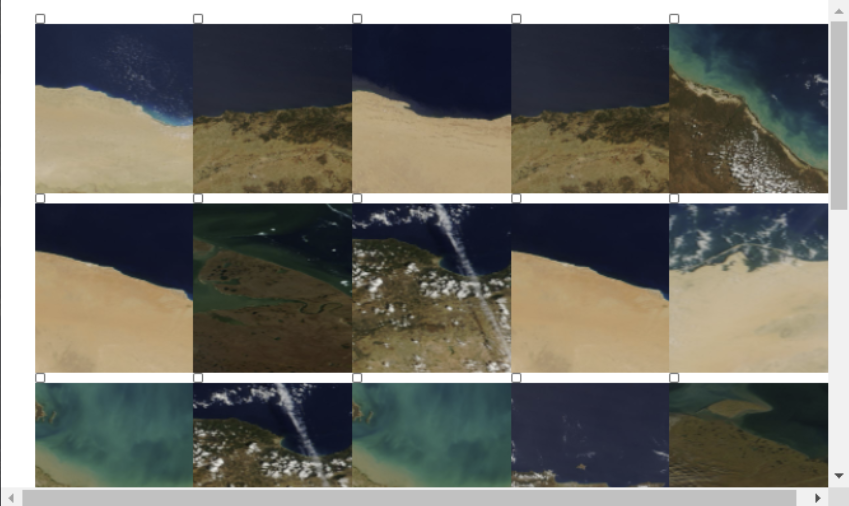}
    \caption{Step 4: The system returns a refined search returned}
    \label{fig:step1}
\end{figure}

\section{Experiments}

Originally, we used a standard ResNet \cite{resnet} model as the featurizer. This had the downside of having a 2048 feature vector size, but also was deficient in detecting the different classes of satellite imagery. The standard ResNet model had high accuracy at clustering images of similar color together, but had limited capabilities beyond that. So, we took a ResNet model and added a dense layer and retrained it on 10 basic classes (islands, reefs, hurricanes, wildfires, etc.) with the hope of creating a much better featurizer. This is the model we ended up using, and it was able to identify several satellite imagery classes, creating an accurate reverse search engine infrastructure as a whole. Below some test images with this new CNN can be seen with their closest matched images. 
\begin{figure}[htp]
    \centering
    \includegraphics[width=10cm]{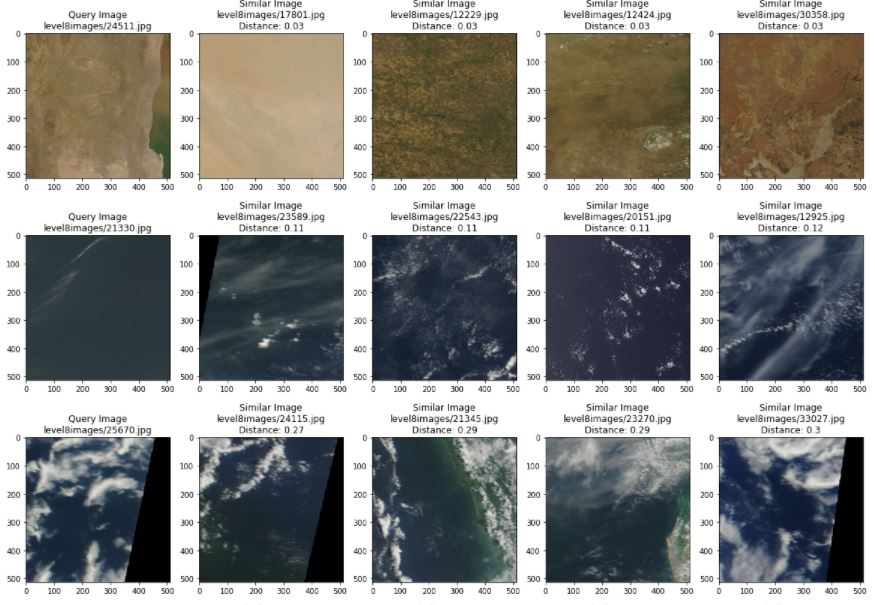}
    \label{fig:hyperplanes}
\end{figure}
\begin{figure}[htp]
    \centering
    \includegraphics[width=10cm]{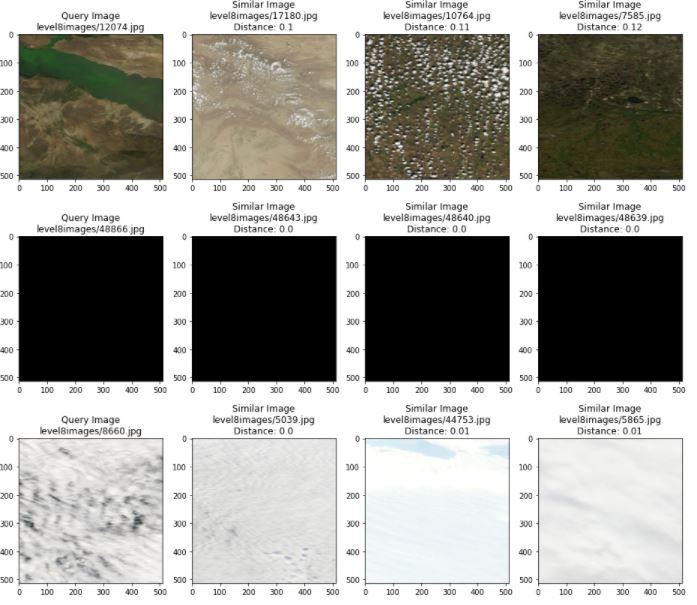}
    \caption{Similar images found to input image with featurization with new model}
    \label{fig:hyperplanes}
\end{figure}

With the above images we can see that this 10-class CNN has demonstrated that a reverse image search over satellite images was possible in an accurate and efficient manner. To expand the scope of the similarity search would be as simple as training the CNN on more classes 

\section{Analysis and Future Work}
By designing this scalable reverse image search system within a cloned version of NASA worldview, we have demonstrated the viability of such a system to exist at a much more massive scale (being able to reverse search more classes of images and have a database of several years of images). The potential of this product to attain such scale is demonstrated by the quick search approximate nearest neighbor search and the extremely scalable storage as a result of the small feature-set required to represent each image in our database. 

In terms of the model used for the reverse image search, the current CNN is only applicable to the classes of labeled images that it was trained on. The issue with this approach is that we cannot possibly train a model on all the possible classes of satellite imagery, since there are too many possible classes. Further, it would be a infeasible to label images to train the model in a supervised manor for reverse image search. For this reason, going forward, a self-supervised model will be sought out to find similarities between images without specific classification and labelling. 

Further, we would like to dockerize the reverse image search architecture so that it could be used by scientists with any dataset and model. Scientists would be required to store their database of images as a collection of the features of all the images (featurized by the custom model) and provide their custom model into this dockerized code as a .h5 file.

\section{Conclusion}
In this paper, we have demonstrated that the creation of a reverse image search system on Worldview is something that is very feasible. Via our compression of search time with approximate-nearest neighbor search, this tool can be used by researchers to very quickly assemble an image database for their research purposes, thus cutting down the threshold to start research significantly. The reverse search engine is able to search over a large database of images because of the 128 length feature vector storage format. This allowed us to speed up search along with shrink the storage of the enormous satellite image databases. We also found that using a standard ResNet \cite{resnet} model was not ideal for reverse-image search, since the ResNet model was good when classifying with respect to color, but lacked finer feature detection. This was improved when we experimented with a CNN of our own that was trained on satellite images, and this was able to much better identify the more granular features in the satellite imagery.

%You may use color figures.  However, it is best for the figure captions and the paper body to be legible if the paper is printed in either black/white or in color.


\begin{thebibliography}{9}

\bibitem{LearningsfromFrontierDevelopmentLab} 
S. Ganju, A. Koul, A. Lavin, J. Veitch-Michaelis, M. Kasam, and J. Parr
 \emph{Learnings from
Frontier Development Lab and SpaceML - AI Accelerators for NASA and ESA}. 
2020.

\bibitem{seeley2020kdf} 
M. Seeley, F. Civilini, N. Srishankar, S. Praveen, A. Koul, A. Berea, and H. El-Askary
 \emph{Knowledge Discovery Framework}.  
2020.

\bibitem{resnet}
K.He X. Zhang S. R.Jian Sun
 \emph{Deep Residual Learning for Image Recognition}.
2016.

\bibitem{ANN}
Wen Li, Ying Zhang, Yifang Sun, Wei Wang, Wenjie Zhang, Xuemin Lin
 \emph{Approximate Nearest Neighbor Search on High
Dimensional Data — Experiments, Analyses, and
Improvement (v1.0)}.
2016.


\end{thebibliography}
\end{document}